%% file: main.tex
\documentclass[11pt, letterpaper]{article}
\usepackage[utf8]{inputenc}
\usepackage[margin=1.5cm]{geometry}
\usepackage{titlesec}
\usepackage{tabu}
\usepackage{longtable}
\usepackage{enumitem}
\usepackage{caption}
\usepackage{amssymb}
\usepackage{xcolor}

\usepackage[hyphens]{url}  
\usepackage{subfigure}
\usepackage{graphicx} 
\usepackage{bm}
\usepackage{bbm}
\usepackage{color}
\usepackage{amsfonts}
\usepackage{stfloats}
\usepackage{tikz}
\usepackage{multirow}
\usepackage{amsmath}
\usepackage{amssymb}

\newlist{selectlist}{itemize}{2}
\setlist[selectlist]{label=$\square$,leftmargin=*,noitemsep,topsep=0pt}

\usepackage{lmodern}

\usepackage{hyperref}
\hypersetup{
    colorlinks=true,
    linkcolor=blue,
    filecolor=magenta,      
    urlcolor=blue,
}
 
\titleformat{\section}[block]{\hspace{1em}\bfseries}{\thesection.}{0.5em}{} 
\titleformat{\subsection}[block]{\hspace{1em}}{\thesubsection}{0.5em}{}

\begin{document}
\begin{flushleft}

\textbf{Article information}\\
\vskip 0.5cm
\textbf{Article title}\\
A Multilingual Dataset of COVID-19 Vaccination Attitudes on Twitter
\vskip 0.5cm
\textbf{Authors}\\
Ninghan Chen\textsuperscript{\rm 1}, Xihui Chen*\textsuperscript{\rm 1}, 
Jun Pang\textsuperscript{\rm 1, 2}
\vskip 0.5cm
\textbf{Affiliations}\\
 \textsuperscript{\rm 1} Faculty of Sciences, Technology and Medicine,  University of Luxembourg, L-4364 Esch-sur-Alzette, Luxembourg\\
    \textsuperscript{\rm 2} 
    Interdisciplinary Centre for Security, Reliability and Trust,  University of Luxembourg, L-4364 Esch-sur-Alzette, Luxembourg
    
\vskip 0.5cm
\textbf{Corresponding author’s email address and Twitter handle}\\
\{ninghan.chen, xihui.chen, jun.pang\}@uni.lu
\vskip 0.5cm
\textbf{Keywords}\\
COVID, vaccination, attitude, Twitter, dataset, health
\vskip 0.5cm
\textbf{Abstract}\\ 
Vaccine hesitancy is considered as one main cause of the stagnant uptake ratio of COVID-19 
vaccines in Europe and the US where vaccines are sufficiently supplied. 
Fast and accurate grasp of public attitudes toward vaccination is critical to 
address vaccine hesitancy, and 
social media platforms have proved to be an effective source of public opinions. 
In this paper, we describe the collection and release of a dataset of tweets related to 
COVID-19 vaccines. 
This dataset consists of the IDs of 2,198,090 tweets collected from Western Europe, 
17,934 of which are annotated with the originators' vaccination stances. 
Our annotation will facilitate using and developing data-driven models to
extract vaccination attitudes from social media posts and thus further 
confirm the power of social media in public health surveillance.   
To lay the groundwork for future research, 
we not only perform statistical analysis and visualisation of our dataset,
but also evaluate and compare the performance of established text-based benchmarks 
in vaccination stance extraction.  
We demonstrate one potential use of our data in practice
in tracking the temporal changes of public COVID-19 vaccination attitudes. 

\vskip 0.5cm
\textbf{Specifications table}\\
\vskip 0.2cm 
%
%
%
\begin{longtable}{|p{33mm}|p{124mm}|}
\hline
\textbf{Subject}                & Social science, Data Science, Computer Science\\
\hline                         
\textbf{Specific subject area}  & Vaccination attitudes, Sentient analysis, Social Media\\
\hline
\textbf{Type of data}           & Text 
\\                                   
\hline
\textbf{How the data were acquired} & The data are collected through the official 
Twitter APIs based on effective keywords commonly used in related datasets. 
The annotation is conducted by a group of 10 multilingual volunteers with three rounds.
\\
\hline                         
\textbf{Data format}            & 
                         Raw (Primary)\newline
                         Labelled (Secondary)
\\                                                    
\hline
\textbf{Description of          
data collection}             & 
The data collection contains the IDs of 2,198,090 tweets related to COVID-19 
vaccination which are posted by active Twitter users residing in 
four geographically adjacent countries in Western Europe. 
The collected tweets are involved in online discourses related to COVID vaccines
or vaccination covering a period of about 14 months after the outbreak of the pandemic. 
The tweets are filtered according to a list of keywords that are commonly adopted 
in previous data collections related to vaccination (not limited to COVID-19) 
and prove to ensure good coverage ratios. 
To enable the use of data-driven methods based on machine learning and deep 
learning, 17,934 tweets are annotated with affective vaccination stances 
(i.e., positive, negative and neutral). The annotation is conducted in three rounds 
with 10 multilingual students among which a high-level agreement is achieved.

\vskip 0.2cm
\\                         
\hline                         
\textbf{Data source location}   & 
\noindent
\textit{ $\bullet$ Country:} France, Germany, Belgium, Luxembourg\\
\hline                         
\hypertarget{target1}
{\textbf{Data accessibility}}   & 

         		Repository name: Zenodo \newline
                         Data identification number: 10.5281/zenodo.5851407\newline
                         Direct URL to data: \url{https://doi.org/10.5281/zenodo.5851407}

\\                         
\hline                         
%
%
%
%
%
\end{longtable}


\textbf{Value of the Data}\\
\begin{itemize}
\itemsep=0pt
\parsep=0pt
\item[$\bullet$]
Our dataset contains the IDs of 2,198,090 tweets relevant to the discourses on COVID-19
vaccination for about 14 months after the onset of the pandemic.  
Its large scale and long-time span allow researchers 
to study the vaccination attitude evolution in Western Europe 
before and after the first COVID-19 vaccine was approved. 
\item[$\bullet$] The over 17,000 tweets annotated with vaccination attitudes 
facilitate developing and validating new data-driven methods, e.g., in  
Natural Language Processing (NLP), to extract vaccination attitudes 
and other subjective opinions from social media posts.
\item[$\bullet$] The validated performance of existing text-based NLP methods 
for opinion extraction demonstrates the power of social media in 
tracking fine-grained temporal changes of vaccination attitudes on a daily or 
weekly basis. The tracking can subsequently lead to timely and proactive interventions 
in this fast-developing pandemic or future public health events of similar types. 
\item[$\bullet$] The multilingualism of our annotated tweets provides a reliable data 
source to evaluate existing and new language transformers in dealing with multilingualism.

\end{itemize}
\vskip 0.5cm

\textbf{Data Description}\\
\vskip 0.5cm


We released the IDs of 2,198,090 tweets related to COVID-19 vaccines from 54,381 active 
Twitter users  between January 20, 2020 and March 15, 2021, spanning about 14 months. 
The tweets are originated from users located in four adjacent Western European countries which are among 
the first group of regions that received and administered COVID-19 vaccines 
and are being hit badly by the pandemic: Belgium, Germany, France and Luxembourg. 
We manually annotated 17,934 tweets with affective vaccination stances 
(i.e., positive, negative and neutral). 
Due to IRB review and the Twitter Terms of Service,
only tweet IDs are published. All tweets are publicly accessible and 
researchers are recommended to download them with official Twitter APIs.

Although social media posts have been used to study vaccination attitudes 
since the outbreak of the pandemic~\cite{LL21}, only a few datasets are publicly available. 
Pierri et al.~\cite{pierri2021vaccinitaly} published a dataset collected from Twitter and Facebook
recording Italian users' discussions about vaccination. 
DeVerna et al.~\cite{DeVerna0TBALTYM21}
released the CoVaxxy dataset composed of 
English-language tweets about the COVID-19 vaccination generated from the US.  
Chen et al.~\cite{chen2021mmcovar} published the MMCoVaR dataset which contains only 24,184 tweets
related to COVID-19 vaccines, spanning less than one month. 
%
Our dataset differs from the above datasets from three aspects. 
First, our released tweets covers a sufficient amount of time before and after the 
administration of the first COVID-19 vaccine. 
This enables studies on the evolution of public vaccination attitudes. 
Second, the tweets are originated from four countries which can well 
portray the first group of COVID-19 vaccine receivers and other European countries 
hit badly by the pandemic. 
Last not least, our annotated tweets allow for utilising established data-driven 
methods to learn vaccination attitudes from tweets, and facilitate the development 
of new methods. 

Our release consists of two files: \texttt{all\_tweet} and \texttt{annotated\_tweet}.
The former lists the IDs of the 2,198,090 collected tweets while the latter
is composed of a table of 17,934 rows with two columns: \emph{tweet\_id} and \emph{label}.
Each row corresponds to a tweet identified with the \emph{tweet\_id} and the \emph{label}
field gives its affective attitude towards vaccination. Specifically, we have five labels 
which are described as follows:
\begin{itemize}
\item \textbf{Positive (PO)}: The originator expresses his/her support for the vaccines or vaccination 
in the sense that vaccine or vaccination can effectively protect the public, and will be or 
has been vaccinated. 
\item  \textbf{Negative (NG)}: The originator expresses doubts or disbelief about the effectiveness 
of the vaccines or vaccination in combating the pandemic, or hesitates or refuses to be vaccinated.
\item \textbf{Neutral (NE)}: No explicit attitude or intention are expressed.
\item  \textbf{Positive but dissatisfaction (PD)}: The originator expresses dissatisfaction or 
complaints about the current policies or measures against COVID-19, but still holds a positive attitude 
towards vaccination.
\item \textbf{Off-topic (OT)}: The content is irrelevant to COVID-19 vaccines or vaccination.
\end{itemize}
We give some examples for each label in Table~\ref{tab:samples}.
Special attention should be paid to the label PD. We notice that there exist a large portion of tweets
expressing the originators' negative feelings or disagreement about the way 
the governments handled the pandemic such as 
complaints about lock-downs. However, the originators still showed their belief in 
vaccines as an effective and ultimate measure to beat the virus. 
Such tweets use terms which are negative inherently  
and if not explicitly separated, they will confuse classification methods 
with the ones that should be labelled as NG. 
\begin{table}[h]
\centering
\caption{Tweet examples.} 
\label{tab:samples}

\resizebox{1\linewidth}{!}
{ \begin{tabular}{|l|l|}
\hline
\textbf{Label} & \textbf{Example (Translated to English)}                    \\ \hline \hline
{\textbf{PO}} &
  {We have a new weapon against the virus: the vaccine. Hold together, again.} \\\hline
\textbf{NG} &
  \begin{tabular}[c]{@{}l@{}}My daughter, a nurse at the AP-HP, on the vaccine "Ah ah ah! They don't even \\ dream about it, they start with the old ones so that we can attribute the side effects to age".\end{tabular} \\ \hline
\multirow{3}{*}{\textbf{PD}} &
  It’s bad enough for individuals to refuse \#COVID19 \#vaccines for themselves. \\
  & But forcing a mass vax site to shutdown, knowing it means vaccines may go to waste, is criminal. \\
  &Call it pandemicide. \\ \hline
\textbf{NE}      & Have any diabetics been vaccinated? I need some information \\ \hline
\textbf{OT}    & a 10\% discount on pet vaccinations next week.              \\ \hline
\end{tabular}
}
\end{table}

In Figure~\ref{fig:temporal}, we display the number of collected tweets on a daily basis. 
Sudden surges of daily posts can help understand the changes of public attention and 
investigate possible events that cause the changes. 
We can see that COVID-19 vaccines or vaccination were rarely discussed before November 9, 2020
when a sudden surge occurred. Since then, the discussion around them remains popular. 
To understand the events that promoted the increase, 
we extracted the tweets that were generated in the 3 days after November 9, 2020 and 
created  a word cloud to identify the frequently used words (see Figure~\ref{fig:word_cloud}).
With the highlighted words, we infer that the event is probable due to the first 
release of the efficacy of the Prizer/Biotech vaccine. 
With a check on the news, we confirmed our inference. 
Pfizer announced the 90\% efficacy of its vaccine co-developed with 
Biotech.\footnote{\url{https://www.pfizer.com/news/press-release/press-release-detail/pfizer-and-biontech-announce-vaccine-candidate-against}}
With the similar approach, we checked the peak on December 27, 2021 and the 
sudden increase on March 15, 2021. We discovered that the former tweet
peak is attributed to the start of the EU Mass Vaccination Campaign 
while the later increase is likely to be caused by the news that 
France and Germany joined other European countries to temporarily halt the use of the 
Oxford-AstraZeneca vaccine.

\begin{figure}[h]
\begin{minipage}{.61\textwidth}
\centering
\resizebox{1\linewidth}{!}{
\includegraphics[scale=0.55]{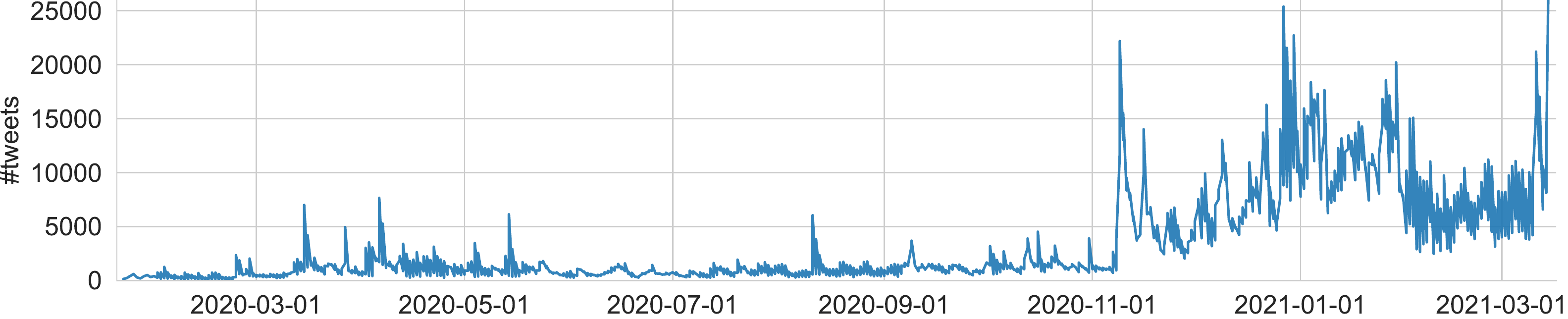}}
\caption{The temporal distribution of tweets on a daily basis.}
\label{fig:temporal}
\end{minipage}%
\begin{minipage}{.38\textwidth}
  \centering
\resizebox{0.8\linewidth}{!}{
  \includegraphics[scale=0.25]{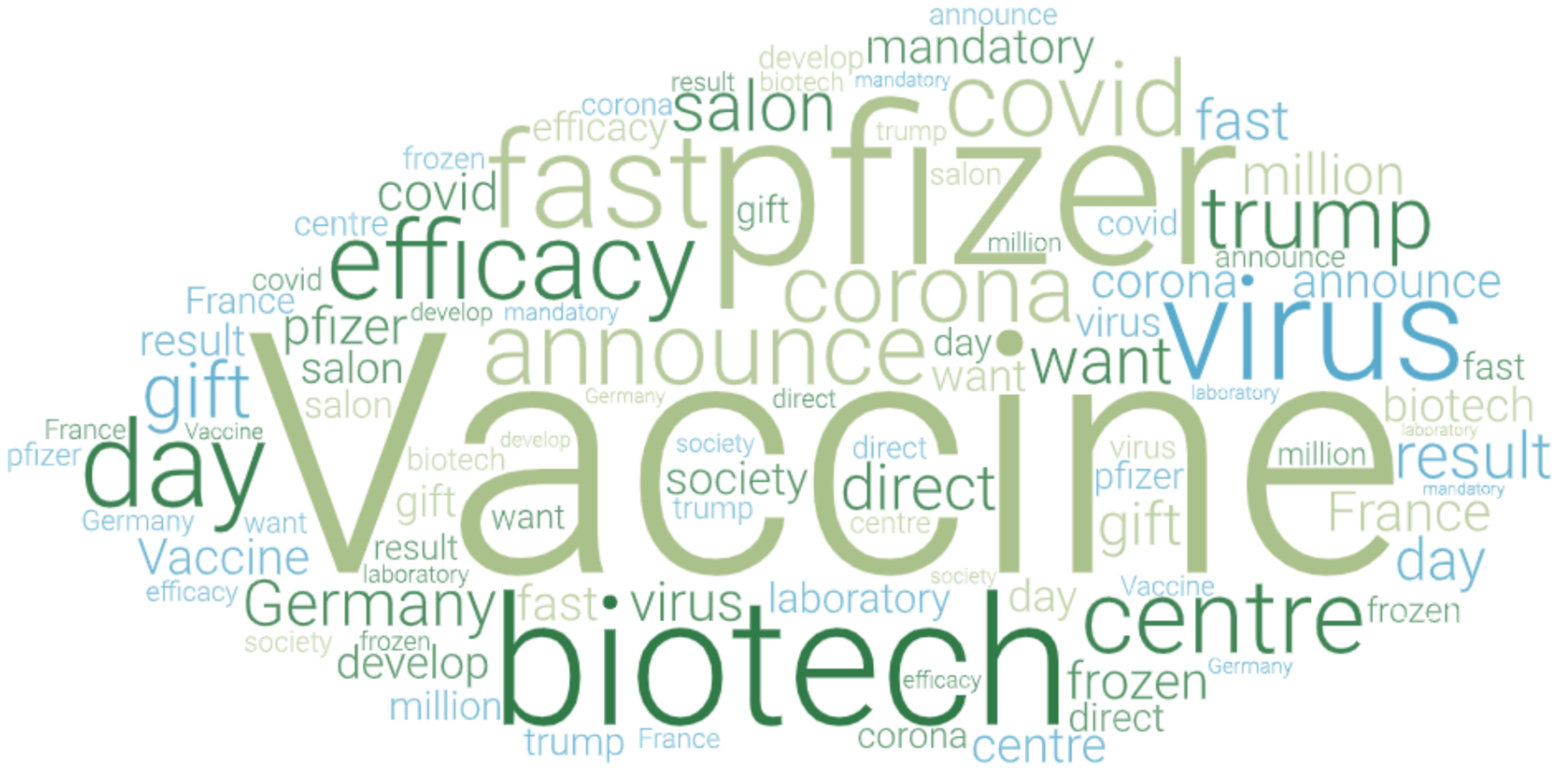}
 }
  \caption{Word cloud of the tweets posted between 11/9/2020 and 11/11/2020.}
  \label{fig:word_cloud}
\end{minipage}%
\vspace{-4mm}
\end{figure}

A character of our tweet dataset is  the multilingualism which is inherent in Europe. 
In Figure~\ref{fig:language_all}, we show the percentages of tweets in the top 
5 most used languages. As the official language of France, Belgium and Luxembourg, 
French is the dominant language which is used in more than 60\% of the collected
tweets. Multilingualism is considered as a challenge in NLP to 
extract subjective opinions from texts. Researchers will benefit from 
our tweet dataset and our annotation in developing and validating new 
NLP methods to address this challenge.
We depict the distribution of annotated tweets over the vaccination attitude labels 
in Figure~\ref{fig:annotated}. 
Adding up those labelled both PO and PD, we can see that more than 60\% of the tweets  express
a positive attitude toward vaccination while about 20\% are associated with negative attitudes. 
\begin{figure}[h]
\centering
\begin{minipage}{.45\textwidth}
  \centering
  \includegraphics[scale=0.45]{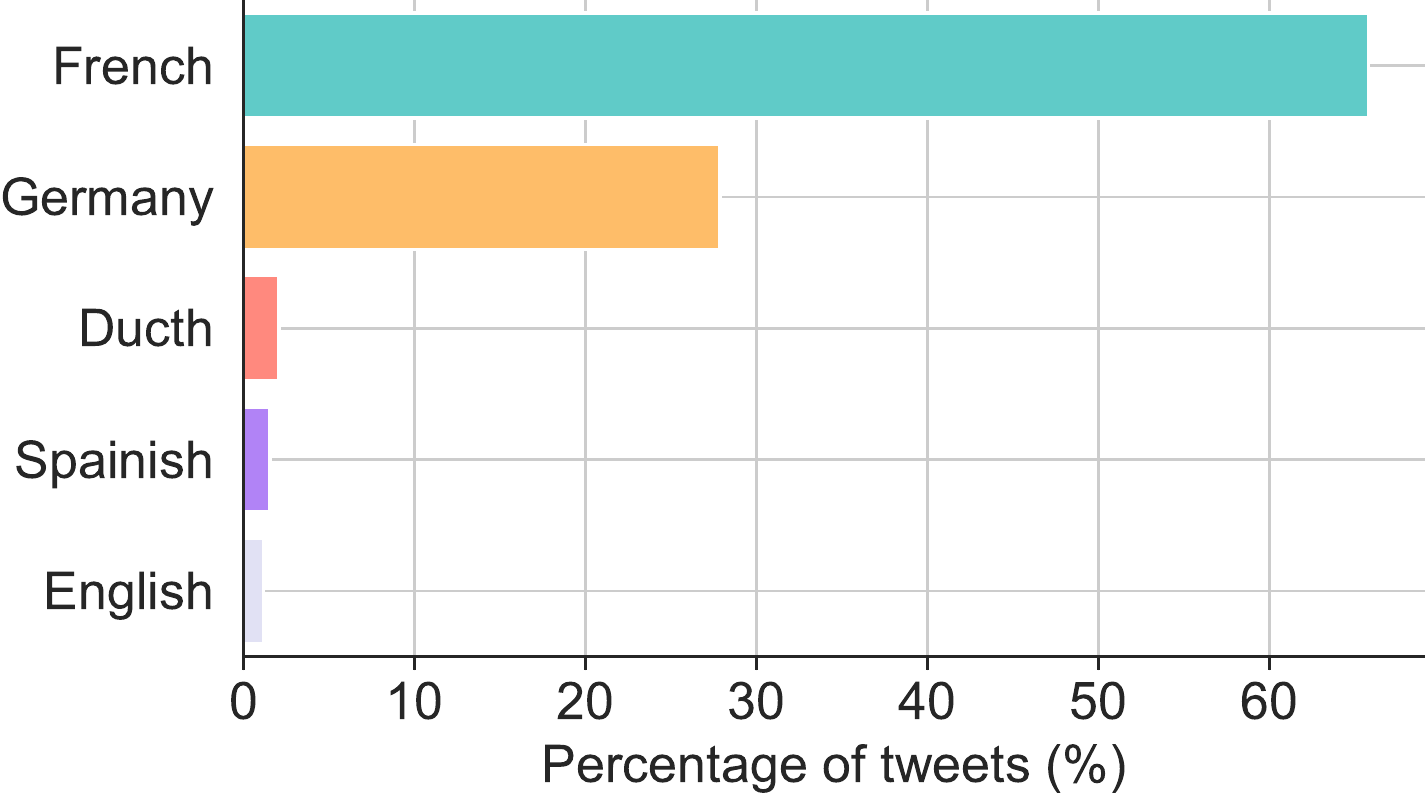}
\caption{Distribution of tweet over languages.}
\label{fig:language_all}
\end{minipage}
\begin{minipage}{.45\textwidth}
  \centering
  \includegraphics[scale=0.45]{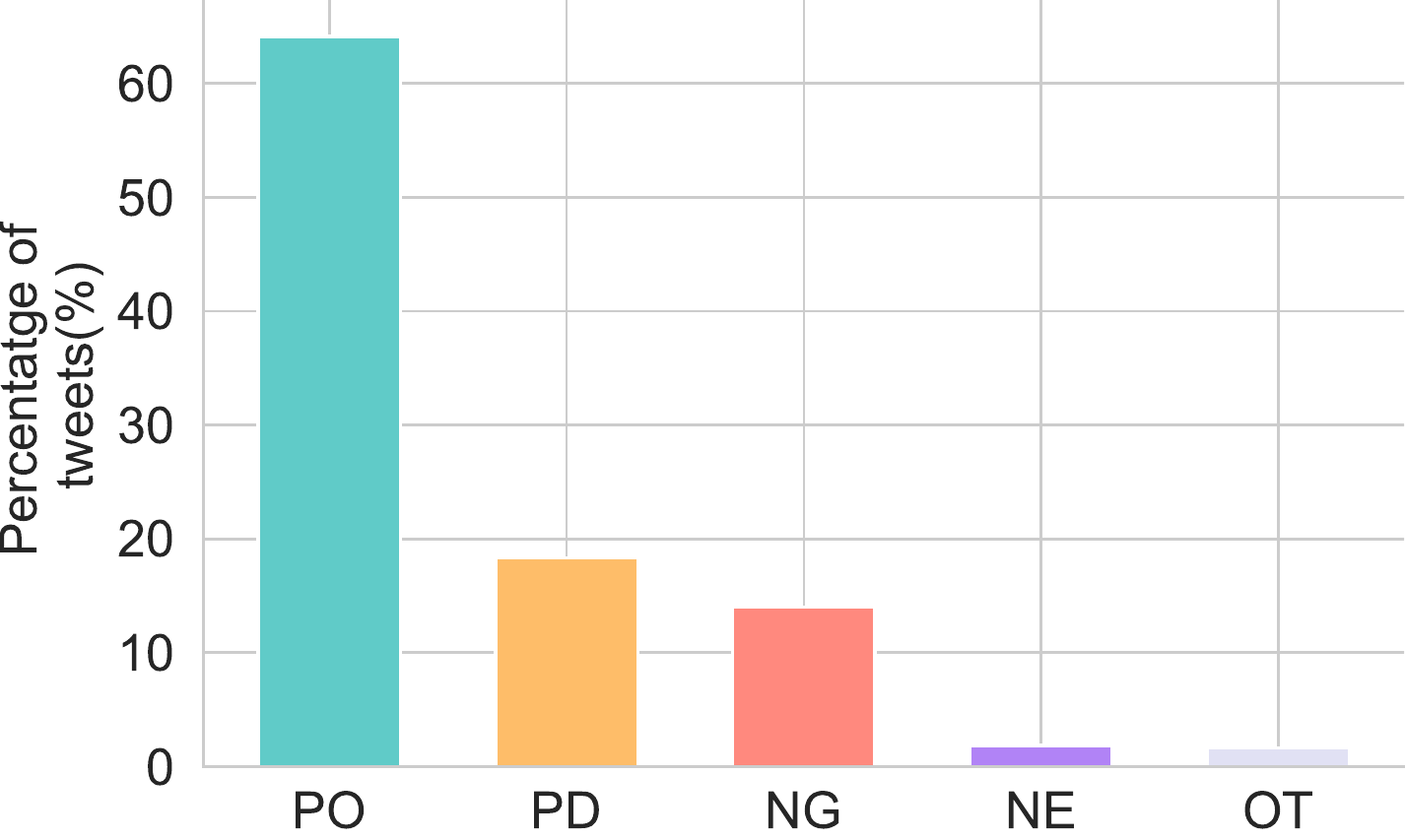}
\caption{Annotated attitude distribution.}
\label{fig:annotated}
\end{minipage}%
\end{figure}


\vskip 0.5cm

\textbf{Experimental design, materials and methods}\\


\section{Data Collection}
\label{sec:Collection}
Our data collection process comprises two steps. 
In the first step, we constructed the set of 
active users from our targeted regions, i.e., Belgium, Germany, France and Luxembourg. 
We say a user is \emph{active} if he/she frequently participates in 
vaccine-related discourse  and interacts with others.
We focused on these users because active users are more likely to express 
their true opinions through tweets. 
In the second step, we streamed tweets from the identified active 
users using manually chosen keywords. 

\subsection{Active user identification}
According to the Twitter policy, the only way to get Twitter users is from the meta-information 
associated with tweets. 
Instead of direct searching tweets, we refer to a publicly available 
tweet dataset~\cite{COVID-19Dataset}. This continuously updated dataset contains 
IDs of tweets related to COVID-19 and originated from users all over the world.
Our data collection spans from January 22, 2020 to March 15, 2021, 
and covers a period from the early stage of the pandemic to the beginning of 
the vaccination campaign.
We downloaded the tweet IDs from the dataset which fall in this period, 
and then crawled the corresponding tweets with the Twitter API.
In total, we downloaded 51,966,639 tweets from which 15,551,266 Twitter users 
are obtained.
To retain users that are active and located in the targeted countries, 
we conducted two sequential filtering operations: 
\emph{location-based filtering} and \emph{activity-based filtering}. 

\smallskip\noindent\textit{\textbf{Location-based filtering.}}
We made use of the geographic information in the metadata of tweets 
to learn users' originating countries. 
If a user has multiple tweets generated from different countries, although this rarely occurs 
among our collected users, we use the country reported in the earliest tweet.
The metadata of a tweet has two fields to store the originator's locations: 
\emph{Geo} and \emph{Place}. 
The \emph{Geo} field records the location generated by the user's device while 
the \emph{Place} field stores the geographic information provided by users.
The \emph{Geo} information is accurate and in a unified format that can be automatically parsed.
However, only about 2\% of tweets come with \emph{Geo} values. 
As a result, we use the \emph{Place} field which is usually ambiguous 
when the \emph{Geo} field is not available.
To regularise the location format and remove the ambiguity, we leveraged the 
the ArcGis Geocoding, which is widely used in previous research 
for the same purpose~\cite{hecht2011tweets}.
For example, the \emph{Place} field value \emph{Moselle}, input by a user, is 
converted into a machine-resolvable location including city, state and country: 
\emph{Mosselle, Lorraine, France}.
Among the selected users, more than 70\% have at least one geographical field filled.
We removed the users located outside the four countries and finally obtained 
767,583 users.

\smallskip\noindent\textit{\textbf{Activity-based filtering.}}
We built a \emph{user interaction graph} to screen out inactive users. 
A user interaction graph is un-directed and weighted. An edge is created 
between two users, e.g., $u$ and  $u'$, when user $u$ retweets or mentions a tweet generated from 
user $u'$ or the other way around. 
The weight of the edge is the number of
retweets or mentions between two end users. 
We first eliminated all edges with weights less than two to exclude the occasional 
interactions. Then we removed the vertices with degrees smaller than 2 to ensure 
active users possess frequent interactions with multiple users. 
In the end, we retained in total 54,381 active users.

\subsection{Vaccine-related tweet streaming}
With the set of active users in the targeted countries, 
in this step, we crawled the tweets originated or retweeted by these users that are 
related to COVID-19 vaccines or vaccination.
Same as previous studies~\cite{yousefinaghani2021analysis,DeVerna0TBALTYM21}, we use keywords 
related to COVID-19 vaccines  to filter tweets. 
Two methods have been adopted to select vaccine-related keywords in the literature~\cite{DeVerna0TBALTYM21}.  
One is called \emph{snowball sampling} which iteratively enriches the initial set of 
keywords according to the newly downloaded messages. 
The other method directly constructs the set of keywords based on expert knowledge 
and contexts. As many keyword lists are publicly available and produce rather 
good coverage~\cite{hussain2021artificial,schmidt2018polarization}, 
we decided to refer to them and only selected the 
ones with the best coverage to keep the list short.
As the tweets originated from our targeted countries are written in multiple languages, 
which are different from those studied in the previous works, 
we translated the selected keywords when necessary. 
After multiple rounds of manual validation, we used all the keywords 
\emph{\textbf{containing}} the following words as substrings: 
`\emph{vax}', `\emph{vaccin}', `\emph{covidvic}', `\emph{impfstoff}', 
`\emph{vacin}',  `\emph{vacuna}',  `\emph{impfung}'. 
`\emph{sputnikv}', `\emph{astrazeneca}', `\emph{sinovac}',
`\emph{pfizer}', `\emph{moderna}', `\emph{janssen}',  `\emph{johnson}' and  `\emph{biontech}'. 

We used the Twitter Academic Research API to search relevant tweets based on 
the active users' IDs and the keywords. The API allows for downloading at most 500
tweets for each downloading request. In order to ensure a good coverage, 
we constructed a request for each user every month. This enables us to obtain an 
acceptable coverage due to the small likelihood that a user posts more than 500 
tweets related to COVID-19 vaccines in one month.  
In total, we downloaded 2,198,090 tweets whose IDs are released.

\section{Data  Annotation}
\label{sec:Annotation}

\subsection{Annotator training and consolidation}
Since the number of downloaded tweets exceeds our capacity to annotate, 
we selected a number of tweets that can well represent
the linguistic features of COVID-10 vaccination related tweets.
Specifically, we first sorted the downloaded tweets in the descending order 
by their numbers of times being retweeted. 
We then removed the most frequently retweeted tweet from the ordered list and added it to the list of 
tweets to annotate iteratively 
until every active user has at least one posted or retweeted message to annotate. 
In total, we selected 17,934 tweets. 

We hired 10 bachelor students to manually annotate the sampled tweets. 
All annotators are proficient in at least two of the four official languages of the
targeted countries, and, in the meantime, were active on Twitter.
One author of this paper acted as the coordinator in charge of annotator training and 
annotation consolidation.
Each annotator received a tutorial from the coordinator explaining the semantics of 
all labels with examples. We also distributed a guideline illustrating the 
workflow on the Doccano platform\footnote{github.com/doccano/doccano} we built to collect annotators' input. 
To ensure that all annotators held the correct understanding, 
we conducted a pilot annotation process 
in which all annotators were first asked to annotate 100 tweets. 
The coordinator verified their annotations and provided additional explanations if necessary. 
We repeated the process with another 100 tweets.
After two rounds of training, annotators succeeded in understanding the labels and 
also became familiar with the Doccano platform.

We first selected one annotator to annotate all the tweets and this full annotation  
took approximately 60 hours. We then randomly assigned to each of the rest 9 annotators 
around 2,000 tweets and asked them to validate the labels. 
Meanwhile, the coordinator went through all annotated tweets.
When disagreeing with the labels given by the first annotator, they added 
new labels. This validation took the coordinator about 60 hours and each of the other annotators 4 hours.
In this way, our annotation strategy ensured each tweet be labelled three times. 
To solve the conflicts, the coordinator consolidated all annotations.
The label agreed by at least two annotators is set as the final annotation. 
For those with three different labels, 
the coordinator communicated with 
the other two annotators and  picked the most appropriate one.

\subsection{Annotator agreement}
To ensure the quality of our annotations, we leverage three widely accepted 
measurements to quantitatively evaluate the inter-annotator reliability for each label:
Average Observed Agreement (AOA)~\cite{fleiss2013statistical}, 
Fleiss' kappa~\cite{fleiss2013statistical}, and 
Krippendorff's Alpha~\cite{krippendorff1970estimating}. 
AOA is the average observed agreement between any pair of annotators.
The term ``observed agreement" in AOA refers to the proportion of labels two annotators agree with.
Both Fleiss' kappa and  Krippendorff's Alpha are applicable to measure the agreement 
between a fixed number of annotators, 
with the difference that Krippendorff's Alpha can handle missing labels. 
The values of all the three measurements range from 0 to 1, 
where 0 indicates complete disagreement and 1 indicates absolute agreement.
For Fleiss' Kappa, 0.41-0.60, 0.61-0.80, and 0.81-1.0 are considered as moderate agreement, 
substantial agreement, and excellent agreement, respectively~\cite{fleiss2013statistical}.
Krippendorff's Alpha is more rigorous than normal standards~\cite{krippendorff1970estimating}. 
Values between 0.667 and 0.800 are deemed acceptable, while values greater than or equal to 0.8 
are considered highly reliable~\cite{krippendorff1970estimating}.

\begin{table}[h]
\caption{Inter-annotator agreement (PO: Positive,
NG: Negative, NE: Neutral, PD: Positive but dissatisfaction, OT: Off-topic).} 
\label{tab:agreement}
\centering
{
{
\begin{tabular}{|l|r|r|r|}
\hline
\textbf{Label} & \multicolumn{1}{l|}{\textbf{AOA}} & \multicolumn{1}{l|}{\textbf{Feliss' kappa}} & \multicolumn{1}{l|}{\textbf{Krippen-dorf's Alpha}} \\ \hline\hline
PO                     & 0.72 & 0.73 & 0.73 \\ \hline
NG                    & 0.82 & 0.88 & 0.88 \\ \hline
NE                    & 0.74 & 0.78 & 0.77 \\ \hline
PD & 0.61 & 0.63 & 0.62 \\ \hline
OT                    & 0.83 & 0.87 & 0.86 \\ \hline
\end{tabular}
}
}
\end{table}

Table~\ref{tab:agreement} summarises the inter-annotator agreement for each annotation label. 
We can see that for all labels, AOA scores range from 0.61 to 0.83. 
This implies that most of the annotations have at least two annotators in agreement.
The values of the other two measurements are close. 
The annotators achieved the highest rank of agreement according to Fleiss' kappa and Krippendorff's Alpha 
for both the labels NE and OT and the second highest rank on labels PO and NE. 
The annotators' agreement on PD falls drastically compared to other labels, but still 
remains moderate according to the ranking criteria of the Fleiss' Kappa measurement. 
This can be explained by our difficulties during annotation in dealing with
the special linguist features of PD tweets, i.e., frequently used negative terms or 
sarcastic expression. 
A closer look will lead to another observation that  
the extent of agreement on the label PO is slightly lower. 
A careful manual investigation reveals that a large proportion of 
disagreed annotations also attribute to the sarcasm and irony made to express their 
opinions about anit-vaccination. 
This identified challenge to handle sarcasm is consistent with the previous finding that 
people frequently are confused by sarcasm, 
which makes comprehension difficult~\cite{reyes2013multidimensional}. 
We take the following tweet as an example: 
\emph{I am very disappointed! 16 days after my first injection of the vaccine against 
\#covid19 I still don't get the 5g.} 
This tweet uses ironic expression joking about anti-vaccination comments but in fact 
delivers definite supporting attitude for vaccination. 
Such tweets produced misunderstandings among annotators, 
which are solved in our consolidation phase.

\section{Vaccination attitude calculation with NLP}
The application of deep learning and machine learning has revolutionised  
NLP, especially in extracting opinions or sentiments from textual contents. 
Compared to machine learning models which rely on manually constructed features, 
deep learning models can learn effective features automatically 
with little manual intervention.  
Extensive empirical evidence has proved the overwhelming performance
of deep learning models in NLP studies~\cite{chakravarthi2020sentiment}.  
We trained and ran several well established NLP models based on machine 
learning and deep learning with our annotated tweets. 
A good performance of the trained models in classifying tweets 
will attest the utility and trustworthiness of our annotation. 

\smallskip\noindent
\textbf{Experiment setup.}
We select Random Forest (RF) and Support Vector Machines (SVM) as the representative machine learning 
models due to their wide use. Regarding deep learning models, we use BERT~\cite{DCLT19}, 
RoBERTa~\cite{LOGDJCLLZS19}, and  DistilBERT~\cite{abs-1910-01108}. Such deep learning models are pretrained and produce 
a low-dimensional representation for any given piece of text, which can be used as input for 
downstream classification methods.

We preprocess the tweets by removing mentions of other users with `@’, 
quoted hyperlinks and `RT’ which stands for ``retweet''. 
We remove tweets with the label `off-topic' due to their small proportion.
To test the models' capability in dealing with multilingualism, 
we construct three datasets: the original annotated tweets, 
the French-language annotated tweets and the German-language annotated tweets.
We divide each of these three datasets into training, testing  
and validation set with the ratio 80\%, 10\% and 10\%, respectively.
To train the selected machine learning models,  we use TfidfVectorizer~\cite{scikit-learn} to 
convert the preprocessed tweets into the bag of $n$-gram vectors. 
We use grid search as the optimiser for SVM and RF.
For deep learning methods, we adopt their publicly available implementation for text embedding  
and keep their default settings. 
The text embeddings are then sent to a fully-connected ReLU layer with dropout. A linear layer is added on the top of the final outputs for regression with softmax as the activation function. 
We use CrossEntropyLoss as the loss function and Adam as the optimiser~\cite{KingmaB14}.  
All models are trained for 30 epochs
for optimisation with the learning rate of $0.00001$, 
and batch size of 32. We set the maximum sequence length as 128, which defines the maximum 
numbers of tokens in a tweet that can be processed.

\smallskip\noindent
\textbf{Result analysis.}    
Table~\ref{tab:Classification} shows the classification performance evaluated with 
conventional measurements. 
First, we can see that the deep learning methods outperform the machine learning methods
in classifying both multilingual tweets and those in single languages, and their
performances are close. This confirms the findings in~\cite{hu2020xtreme}. 
Second, the results show that multilingualism affects the classification 
performance of deep learning models, although we use a pre-trained model 
specifically for classifying tweets in multiple languages. 

By comparing with the models' performance 
on other classifying tasks in the literature~\cite{chakravarthi2020sentiment,hu2020xtreme}, 
we observe that the models can achieve the same-level performances. 
This implies that our annotation is trustworthy and 
useful for future research on vaccination attitude learning.
The results we showed in Table~\ref{tab:Classification} can thus be referred to 
as benchmarks for comparison. 

\input{experimental_results}

\section{Temporal analysis of vaccination attitudes}
We present a description of an application of our tweet 
dataset and annotation to illustrate one potential use of our release.
Specifically, we visualise the temporal evolution of the various vaccination attitudes
and analyse the possible causes to some changes that require more attention. 

We apply XLM-Roberta model trained in the previous experiment to calculate the 
vaccination attitudes of the tweets in the tweet dataset. 
Figure~\ref{fig:pre_annotation} shows the distribution of the predicted 
vaccination attitudes. We can see the distribution is similar to that shown in 
Figure~\ref{fig:annotated} with slightly more tweets with negative attitudes.
\begin{figure}[htb]
\centering
\includegraphics[scale=0.45]{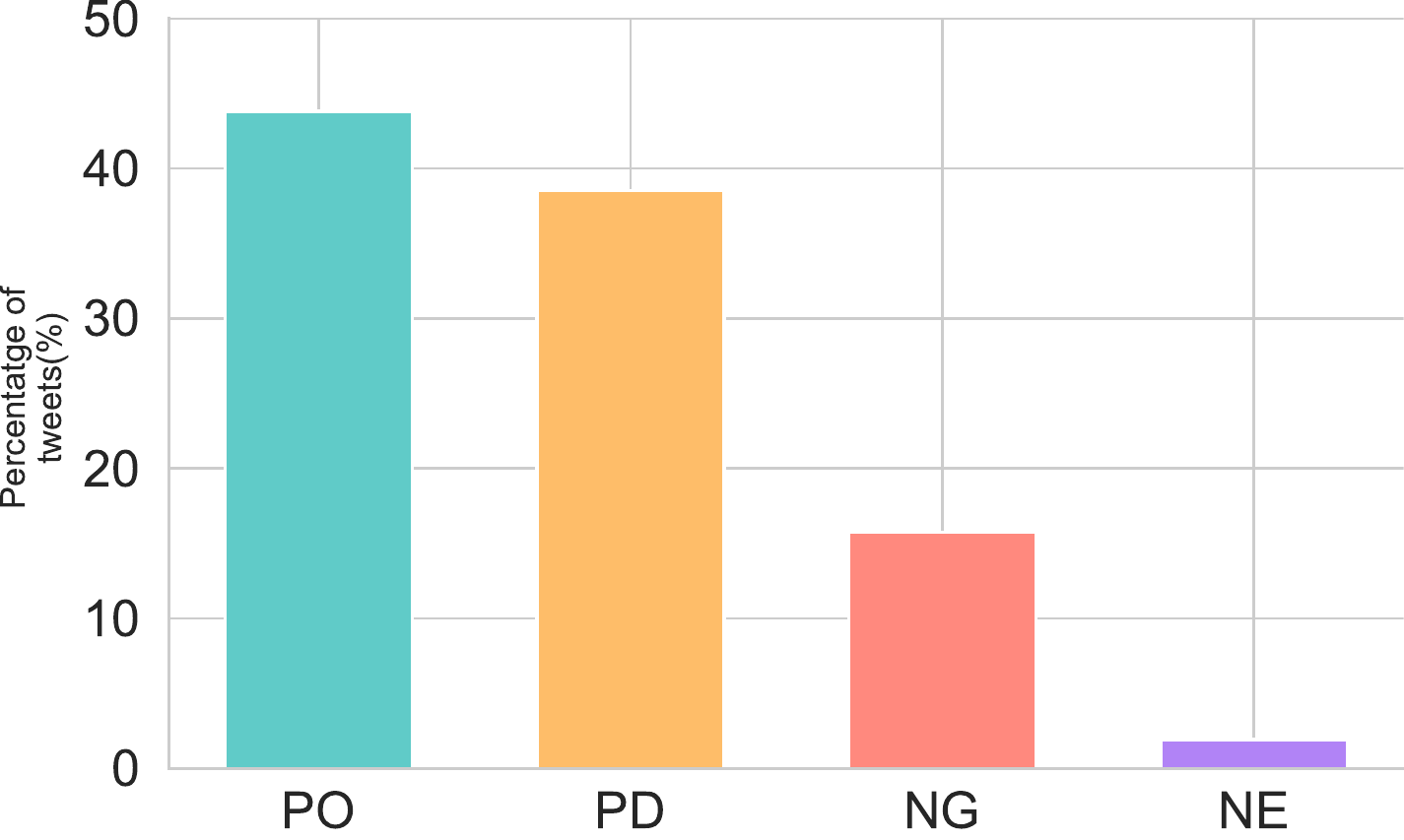}
\caption{Attitude distribution of all tweets.}
\label{fig:pre_annotation}
\vspace{-4mm}
\end{figure}

We draw the temporal evolution of the numbers of tweets that are classified as 
NE, PO and PD in Figure~\ref{fig:pre_temporal}.
Note we ignore the tweets with label OT due to their small amount. 
Recall that significant discussion on COVID-19 vaccines started from November 8, 2020 
as shown in Figure~\ref{fig:temporal}. Therefore, we focus on the 
attitude changes from that date. 
We can see that the number of tweets containing different attitudes toward 
vaccination changes over time. 
Compared to the temporal distribution of daily tweets shown in Figure~\ref{fig:pre_temporal}, 
we observe that the growth in the total number of 
tweets is not accompanied by a proportional change of vaccination attitudes.
Specifically, the number of neutral tweets 
varies less from day to day and remains stable at the same level compared to 
tweets with other attitude labels.  
Based on previous research reporting that the content of tweets 
is highly correlated with real-world situations~\cite{PCHG20}, 
we make a hypothesis that real-world events may contribute to 
the fluctuating numbers of tweets with different attitudes. 

\begin{figure}[tbh]
\centering
\includegraphics[scale=0.45]{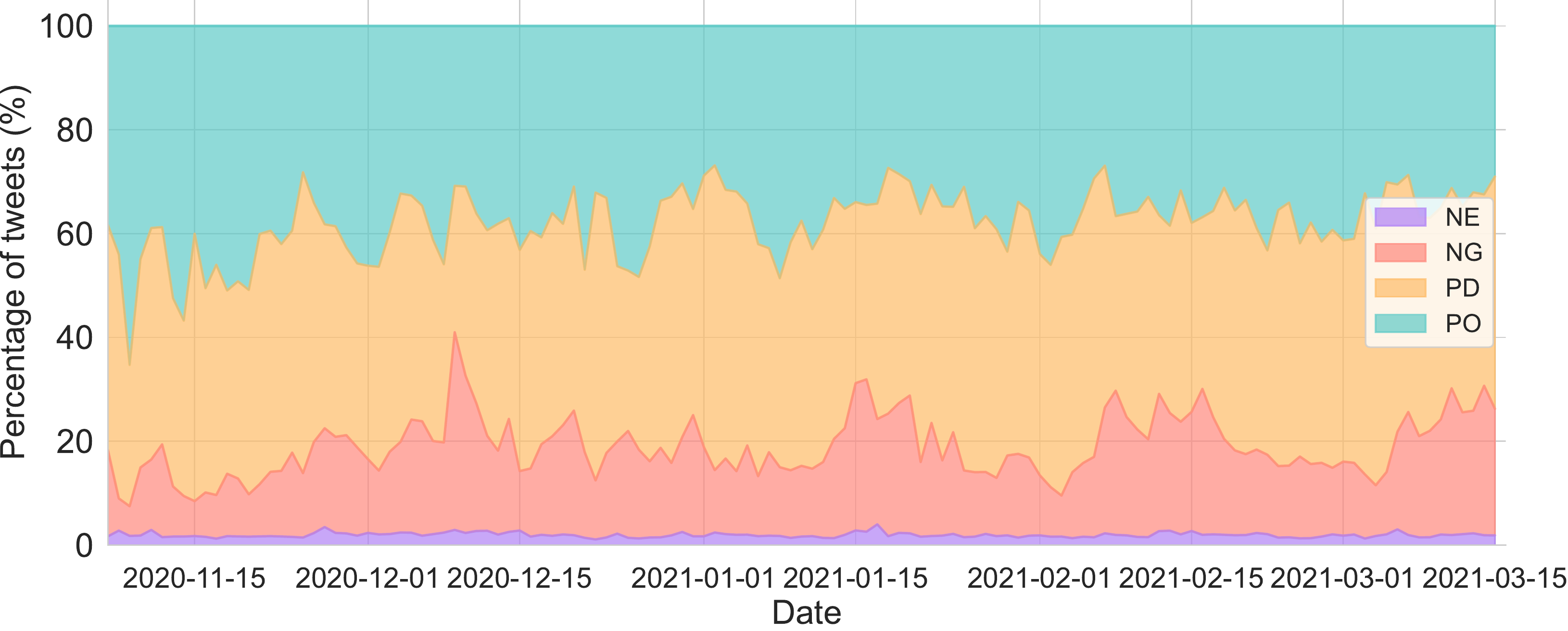}
\caption{Temporal distribution of tweets with different vaccination attitude labels.}
\label{fig:pre_temporal}
\vspace{-4mm}
\end{figure}

In monitoring vaccine hesitancy, more attention should be paid to the fluctuations of 
the negative attitudes. In the following, we take four peaks of the curve of negative tweets 
as examples and discuss the potential causes. We follow the same approach  
used in the temporal analysis of tweets volumes. 
We first plot word clouds to identify the most frequently used keywords and then 
search these keywords on the Internet to understand the events that may contribute to the changes.
The first peak happened around December 9, 2020 due to negative news and misinformation 
about the vaccine efficacy.   
For instance, on December 9, 2020, two healthcare workers were reported to 
have experienced symptoms after receiving their first shots, 
and on December 11, Sanofi and GSK delayed COVID-19 vaccine.
A wide spread piece of misinformation during this period is also identified, 
saying that 6 people had died after vaccinated and another 4 developed Bells Palsy.
Another two peaks occurred around January 15, 2021 and February 10, 2021, respectively. 
We notice that both of these two peaks attribute to the propagation of 
a large volume of misinformation.
Take two pieces of misinformation identified for the peak in January, 2021 as examples. 
One said that on January 14, the Norwegian Medicines Agency reported that a total of 
29 people had suffered side effects, 13 of which were fatal.
The other was about the death of an Indian healthcare worker after receiving Covid-19 vaccines. 
The last peak in March, 2021 is due to the negative news that multiple countries 
decided to suspend the use of the AstraZeneca vaccine due to the reported negative
effects.

From the above discussion, we can see that with NLP deep learning models, public
vaccination attitudes can be extracted. 
When social network data are available,  
we can track almost in real time the changes of vaccination attitudes and understand
the potential causes. This may finally help the governments 
make corresponding effective intervention methods proactively.

\vskip 0.5cm

\textbf{Ethics statements}\\
We strictly adhered to the Twitter Developer Agreement and Policies\footnote{https://developer.twitter.com/en/developer-terms/agreement-and-policy} 
in the collection and distribution of data. 
Our release is also compliant with the EU General Data Protection Regulation (GDPR). 
The released data do not contain any personally identification information. 
Only tweet IDs and annotation are published.

\vskip0.5cm
\noindent
\textbf{CRediT author statement}\\
\vskip0.5cm

\textbf{Ninghan Chen:} Data curation, Visualization, Investigation. 
\textbf{Xihui Chen:}  Conceptualization, Methodology, Writing- Original Draft.  
\textbf{Jun Pang:}  Supervision, Validation, Writing- Reviewing and Editing.

\vskip0.5cm

\textbf{Acknowledgments}\\

\vskip0.3cm
Funding:  This work is supported by Luxembourg’s Fonds National de la Recherche 
via grants COVID-19/2020-1/14700602~(PandemicGR),
PRIDE17/12252781/DRIVEN and PRIDE15/10621687/SPsquared.
\vskip0.3cm

\textbf{Declaration of Competing Interest}\\

\vskip0.3cm
\begin{itemize}
\item[$\text{\rlap{$\checkmark$}}\square$]{The authors declare that they have no known competing financial interests or personal relationships that could have appeared to influence the work reported in this paper.}

\item[$\square$]{The authors declare the following financial interests/personal relationships which may be considered as potential competing interests: }
\end{itemize}
\vskip0.3cm

\newpage
\textbf{References}\\
\bibliographystyle{plain} 
\bibliography{refs}


\end{flushleft}
\end{document}

%% file: experimental_results.tex
\begin{table*}[htb]
\centering
\caption{Classification results for different benchmarks.} 
\label{tab:Classification}

\resizebox{1.0\linewidth}{!}{
{
\begin{tabular}{|l|rrrr|rrrr|rrrr|}
\hline
 &
  \multicolumn{4}{c|}{Mulitilingual} &
  \multicolumn{4}{c|}{French} &
  \multicolumn{4}{c|}{German} \\ \hline
\textbf{Model} &
  \multicolumn{1}{l|}{Precision} &
  \multicolumn{1}{l|}{Recall} &
  \multicolumn{1}{l|}{F1} &
  \multicolumn{1}{l|}{Accuracy} &
  \multicolumn{1}{l|}{Precision} &
  \multicolumn{1}{l|}{Recall} &
  \multicolumn{1}{l|}{F1} &
  \multicolumn{1}{l|}{Accuracy} &
  \multicolumn{1}{l|}{Precision} &
  \multicolumn{1}{l|}{Recall} &
  \multicolumn{1}{l|}{F1} &
  \multicolumn{1}{l|}{Accuracy} \\ \hline
\textbf{RF} &
  \multicolumn{1}{r|}{0.4317} &
  \multicolumn{1}{r|}{0.3219} &
  \multicolumn{1}{r|}{0.4471} &
  0.4749 &
  \multicolumn{1}{r|}{0.5676} &
  \multicolumn{1}{r|}{0.4829} &
  \multicolumn{1}{r|}{0.4754} &
  0.5510 &
  \multicolumn{1}{r|}{0.51481} &
  \multicolumn{1}{r|}{0.4978} &
  \multicolumn{1}{r|}{0.4611} &
  0.4503 \\
\textbf{SVM} &
  \multicolumn{1}{r|}{0.4001} &
  \multicolumn{1}{r|}{0.3816} &
  \multicolumn{1}{r|}{0.4380} &
  0.4263 &
  \multicolumn{1}{r|}{0.5004} &
  \multicolumn{1}{r|}{0.4256} &
  \multicolumn{1}{r|}{0.4141} &
  0.4998 &
  \multicolumn{1}{r|}{0.4954} &
  \multicolumn{1}{r|}{0.4037} &
  \multicolumn{1}{r|}{0.4093} &
  0.4719 \\
\textbf{mBERT} &
  \multicolumn{1}{r|}{0.6622} &
  \multicolumn{1}{r|}{0.5769} &
  \multicolumn{1}{r|}{0.6132} &
  0.6466 &
  \multicolumn{1}{r|}{0.7016} &
  \multicolumn{1}{r|}{0.6933} &
  \multicolumn{1}{r|}{0.7004} &
  0.7184 &
  \multicolumn{1}{r|}{0.6999} &
  \multicolumn{1}{r|}{0.6875} &
  \multicolumn{1}{r|}{0.6919} &
  0.7038 \\
\textbf{XLM-RoBERTa} &
  \multicolumn{1}{r|}{\textbf{0.6801}} &
  \multicolumn{1}{r|}{\textbf{0.5848}} &
  \multicolumn{1}{r|}{0.6271} &
  \textbf{0.6618} &
  \multicolumn{1}{r|}{\textbf{0.7023}} &
  \multicolumn{1}{r|}{\textbf{0.7018}} &
  \multicolumn{1}{r|}{\textbf{0.7145}} &
  \textbf{0.7086} &
  \multicolumn{1}{r|}{\textbf{0.7102}} &
  \multicolumn{1}{r|}{0.6971} &
  \multicolumn{1}{r|}{\textbf{0.7081}} &
  \textbf{0.7079} \\
\textbf{Distil-mBERT} &
  \multicolumn{1}{r|}{0.6768} &
  \multicolumn{1}{r|}{0.5834} &
  \multicolumn{1}{r|}{\textbf{0.6287}} &
  0.6601 &
  \multicolumn{1}{r|}{0.6978} &
  \multicolumn{1}{r|}{0.6916} &
  \multicolumn{1}{r|}{0.7084} &
  0.7065 &
  \multicolumn{1}{r|}{0.7094} &
  \multicolumn{1}{r|}{\textbf{0.7004}} &
  \multicolumn{1}{r|}{0.7068} &
  0.7071 \\ \hline
\end{tabular}
}}
\end{table*}